\documentclass[10pt,twocolumn,letterpaper]{article}

\usepackage{cvpr} 
\usepackage{times}
\usepackage{epsfig}
\usepackage{graphicx}
\usepackage{amsmath}
\usepackage{amssymb}
\usepackage{url}
\usepackage{booktabs}
\usepackage{adjustbox}
\usepackage{multirow, multicol}
\usepackage[table,xcdraw]{xcolor}
\usepackage{pifont}
\usepackage{threeparttable}

\begin{document}

\title{DifFoundMAD: Foundation Models meet Differential Morphing Attack Detection}

\author{Lazaro J. Gonzalez-Soler, André Dörsch, Christian Rathgeb and Christoph Busch \\
da/sec - Biometrics and Security Research Group\\ Darmstadt, Germany \\
{\tt\scriptsize \{lazaro-janier.gonzalez-soler,andre.doersch,christian.rathgeb,christoph.busch\}@h-da.de}\\
}

\maketitle
\thispagestyle{empty}

\begin{abstract}
In this work, we introduce DifFoundMAD, a parameter-efficient D-MAD framework that exploits the generalisation capabilities of vision foundation models (FM) to capture discrepancies between suspected morphs and live capture images. In contrast to conventional D-MAD systems that rely on face recognition embeddings or handcrafted feature differences, DifFoundMAD follows the standard differential paradigm while replacing the underlying representation space with embeddings extracted from FMs. By combining lightweight finetuning with class-balanced optimisation, the proposed method updates only a small subset of parameters while preserving the rich representational priors of the underlying FMs. Extensive cross-database evaluations on standard D-MAD benchmarks\footnote{The approach will also be evaluated on the Bologna Online Evaluation Platform (BOEP) and NIST FATE MORPH.} demonstrate that DifFoundMAD achieves consistent improvements over state-of-the-art systems, particularly at the strict security levels required in operational deployments such as border control: The error rates reported in the current state-of-the-art were reduced from 6.16\% to 2.17\% for high-security levels using DifFoundMAD\footnote{\url{https://github.com/ljsoler/DifFoundMAD}}.

\end{abstract}

\section{Introduction}
\label{sec:intro}

Face recognition has become a cornerstone of modern authentication systems, ranging from smartphone access to border control. As reliance on these technologies continues to grow, ensuring their robustness against increasingly sophisticated attacks is of paramount importance. Among these, morphing attacks (MAs) pose a serious threat, particularly in identity verification processes involving travel documents such as passports.

MAs consist of digitally combining two or more facial images to create a synthetic identity that represents all contributing subjects. Such morphed images can be used to enrol an authorised individual (the accomplice) into a biometric system, enabling an unauthorised person (e.g., a blacklisted subject) to later impersonate this identity without detection. This vulnerability is especially critical in high-security scenarios such as automated border control, where a live capture is compared against a stored passport image. If the stored passport image is morphed, multiple individuals may successfully authenticate under a single identity, thereby compromising the integrity of the system~\cite{Scherhag-FaceMorphingAttacks-TIFS-2020}.

To address this threat, differential morphing attack detection (D-MAD) has been proposed~\cite{Scherhag-FaceMorphingAttacks-TIFS-2020,Domenico-CombIdentityDMAD-ICIAP-2023,DiDomenico-ACIdA-2024,Kessler-MinPairSelectionMAD-2024}. Unlike single morphing attack detection (S-MAD) approaches~\cite{Scherhag-PRNU-TBIOM-2019,Rachalwar-DepthSMAD-IJCB-2023,Caldeira-MADtion-WACV-2025}, which analyse a single image in isolation, D-MAD operates on a pair of images by comparing a trusted live capture with a suspected morph image. This enables detection at both enrolment and verification stages, making D-MAD particularly suitable for deployment in operational environments such as airport eGates~\cite{Papavasileiou-eGates-2025}, where both accuracy and reliability at strict security levels are essential.

Despite these advantages, current deep learning-based D-MAD systems remain limited in their generalisation capability by the scarcity of diverse labelled data and their sensitivity to variations in morphing tools, acquisition conditions, and subject diversity~\cite{BOEP,Ngan-MAD-NIST-2025,Scherhag-FaceMorphingAttacks-TIFS-2020}. Privacy constraints and the difficulty of collecting large-scale, representative facial datasets further exacerbate these issues. Consequently, many existing methods struggle to generalise to unseen morphing techniques and real-world conditions, limiting their applicability in high-security deployments.

Recent advances in vision foundation models (FM) have demonstrated strong cross-domain generalisation capabilities due to large-scale pretraining on diverse datasets~\cite{Radford-CLIP-ICML-2021,Jia-ALIGN-ICML-2021,Kirillov-SAM-ICCV-2023}. These models provide rich and transferable visual representations that can be adapted to downstream tasks with minimal supervision. While FMs have shown promising results in related areas such as presentation attack detection (PAD)~\cite{Ozgur-FoundPAD-WCACV-2025} and S-MAD~\cite{Caldeira-MADtion-WACV-2025}, their integration into the D-MAD framework remains largely unexplored. Given their ability to capture fine-grained visual cues beyond identity-specific information, they represent a promising direction for improving robustness under cross-database and unknown-attack scenarios.

In this work, we introduce DifFoundMAD, a D-MAD framework that leverages FM representations within the standard differential paradigm. Instead of relying on identity embeddings from face recognition systems, we extract high-level representations using pretrained FMs and model the discrepancy between a live capture and a suspected morph through a differential embedding.  Specifically, both images are processed by two instances of the same FM architecture, each independently adapted via lightweight Low-Rank Adaptation (LoRA)~\cite{Hu-LoRA-ICLR-2022}. While the architectural backbone is shared, the parameters are not tied, allowing each branch to specialise in encoding complementary characteristics of the live capture and the suspected morph. The resulting embeddings are then combined into a differential representation, which enables the model to capture subtle inconsistencies indicative of morphing artefacts while preserving the generalisation capability of the pretrained representations. We hypothesise that FM representations should provide complementary information to identity-based embeddings, particularly when morphing artefacts are subtle. 


The primary contributions of this work are as follows:

\begin{itemize}
    \item We propose DifFoundMAD, a D-MAD framework that incorporates FMs to capture complementary information between suspected morphs and live captures. By leveraging pretrained FM representations, the method achieves consistent improvement in generalisation to unknown MAs and acquisition conditions.

    \item We analyse the limitations of classical D-MAD systems based on identity embeddings and observe that such representations, designed to minimise intra-class variability, may suppress subtle artefacts relevant for morph detection. Our experimental results demonstrate that the DifFoundMAD representation captures both morphing artefacts and identity-related cues.

    \item We conduct an ISO/IEC 20059~\cite{ISO-IEC-20059}-compliant evaluation under cross-database and unknown-attack scenarios, demonstrating that DifFoundMAD achieves state-of-the-art detection performance and high robustness under real-world operating conditions by optimising a minimal number of parameters.
\end{itemize}

The remainder of this paper is structured as follows: Sect.~\ref{sec:related_work} reviews relevant D-MAD literature. Sect.~\ref{sec:approach} presents the proposed method. Sect.~\ref{sec:exp_setup} describes the experimental setup. Sect.~\ref{sec:results} reports the results and discussion. Finally, Sect.~\ref{sec:conclusions} concludes the paper and outlines future work.

\section{Related Work}
\label{sec:related_work}

MAs are recognised by NIST as a major threat to facial recognition systems~\cite{Ngan-MAD-NIST-2025}. By combining facial representations from multiple individuals, an attacker can create a composite image that matches more than one identity, enabling accomplices to pass verification undetected~\cite{Scherhag-FaceMorphingAttacks-TIFS-2020}. Although the EU recommends live capture during enrolment~\cite{EU-MADRegulations-2019}, many countries still accept pre-captured printed photos in the application process for identity documents. If these images are morphed, the resulting document (e.g., passport) becomes a long-term vulnerability in border control systems~\cite{Ferrara-FaceDemorphing-TIFS-2018}.

To mitigate the risks posed by MAs, two main detection strategies have been developed: S-MAD and D-MAD. S-MAD analyses a single image, typically a passport photo, to determine whether it has been morphed, thereby preventing compromised samples from entering official databases. In contrast, D-MAD operates at the verification stage by comparing a trusted live capture with a suspected morph, making it particularly suitable for operational environments such as border control.

\subsection{S-MAD Approaches}

Traditional S-MAD methods rely on handcrafted features to capture textural inconsistencies~\cite{Scherhag-MADMultiAlg-ICBEA-2018,Raja-MADBenchmark-TIFS-2020,Dargaud-PCASMAD-WCACV-2023}, image quality degradations~\cite{Debiasi-PRNUVarianceMAD-BTAS-2018,Scherhag-PRNU-TBIOM-2019}, and noise pattern anomalies~\cite{Venkatesh-MADDeepColorResidualNoise-IPTA-2019,Venkatesh-MADResidualNoise-WCACV-2020}. More recent approaches employ deep neural networks (DNNs) to learn discriminative representations for morph detection~\cite{Zhang-GenSMADDeepRep-CVPR-2024,Caldeira-MADtion-WACV-2025,Tapia-SMADFewShot-Neurocomputing-2025}, including FM-based approaches that improve generalisation under limited supervision~\cite{Paulo-FDMAD-ArXiv-2026}. Hybrid methods combining multiple feature extractors and classifiers have also been explored~\cite{Scherhag-MADMultiAlg-ICBEA-2018,Seibold-DeepSMAD-2017}. 

Recent synthesis-based approaches, such as SelfMAD~\cite{Ivanovska-SelfMAD-ArXiv-2025}, leverage synthetic morph artefact generation and self-supervised learning to improve robustness, achieving significant performance gains in cross-morph scenarios.

\subsection{D-MAD Approaches}

D-MAD methods compare a suspected morph and a live capture to detect inconsistencies indicative of morphing. Early approaches focused on texture and gradient analysis~\cite{Raghavendra-TexturalMAD-ISBA-2019,Singh-RobustMADatABCGate-SITIS-2019}, facial landmark deformation~\cite{Scherhag-LandmarkMAD-ICISP-2018,Damer-DMADFacialLandmark-PR-2019}, and multispectral cues~\cite{Raghavendra-MultispectralDMAD-WCACV-2024}. More recent methods incorporate deep learning, including feature-difference classification~\cite{Damer-DMADMultiDetector-FUSION-2019,Scherhag-FaceMorphingAttacks-TIFS-2020,Ibsen-DifferentialAnomalyDetectionForFacialImages-WIFS-2021,Liu-DMADTriplet-BIOSIG-2024} and federated learning strategies~\cite{Robledo-TowardsFL-MAD-2024}. Extensions have also addressed print--scan morphs~\cite{Ngan-MAD-NIST-2025,Tapia-DMADPrintScan-IEEEAccess-2025}, video-based morphing~\cite{Borghi-VMAD-IJCB-2024}, and zero-shot detection using multimodal models~\cite{Shekhawat-ZeroShotDMAD-Arxiv-2025}.

A related line of work focuses on demorphing, aiming to reconstruct the original contributors from a morphed image using landmark-based inversion~\cite{Ferrara-FaceDemorphing-TIFS-2018}, generative adversarial networks~\cite{Peng-FDGAN-IEEEAccess-2019,Shukla-FacialDemorphing-IJCB-2024,Zhang-FaceDemorphing-IP-2025}, or operational demorphing systems~\cite{Ortega-BorderDeMAD-IEEEAccess-2020}. Role-specific approaches such as ACIdA~\cite{DiDomenico-ACIdA-2024} further extend D-MAD by distinguishing between bona fide, accomplice, and attacker roles.

Despite these advances, most D-MAD approaches struggle to generalise, primarily due to the scarcity of large, diverse, labelled datasets~\cite{Ibsen-TetraLoss-FG-2024,Caldeira-MADtion-WACV-2025}. Privacy constraints limit data collection, and existing datasets (see Tab.~\ref{tab:DB}) are restricted in terms of morphing tools, acquisition conditions, and subject diversity. Additionally, many datasets contain bona fide samples (BS) captured under limited temporal and environmental variability. The removal of publicly available datasets such as Idiap Morph~\cite{Sarkar-IdiapMorph-ICASSP-2022} further exacerbates this limitation.

\subsection{Leveraging Foundation Models}

A promising direction to address the generalisation limitations of D-MAD is the use of FMs, which are pretrained on large and diverse datasets and can be adapted to a wide range of vision tasks~\cite{Awais-FM-TPAMI-2025}. These models provide rich and transferable representations, reducing the dependence on domain-specific labelled data.

Since 2021, several FMs have been proposed for visual understanding, including CLIP~\cite{Radford-CLIP-ICML-2021}, ALIGN~\cite{Jia-ALIGN-ICML-2021}, AIMv1/AIMv2~\cite{El-AimV1-ArXiv-2024,Fini-AImv2-CVPR-2025}, and DINO-based architectures~\cite{Oquab-DINOv2-ArXiv-2023,Simeoni-DINOv3-ArXiv-2025}. While some of these models have been explored in biometrics, including PAD~\cite{Ozgur-FoundPAD-WCACV-2025,GonzalezSoler-ZeroFoundationModelsPAD-FG-2025} and S-MAD~\cite{Caldeira-MADtion-WACV-2025}, their integration into the D-MAD setting remains largely unexplored.


Despite their advantages, FMs also pose challenges, including potential biases and domain mismatches from web-scale pretraining, limited suitability for facial analysis (e.g., CLIP~\cite{Radford-CLIP-ICML-2021}), and high computational cost, which may hinder real-time deployment. These limitations highlight the need for careful adaptation in safety-critical applications.

\section{DifFoundMAD}
\label{sec:approach}

\begin{figure}[!t]
    \centering
    \includegraphics[width=0.8\linewidth]{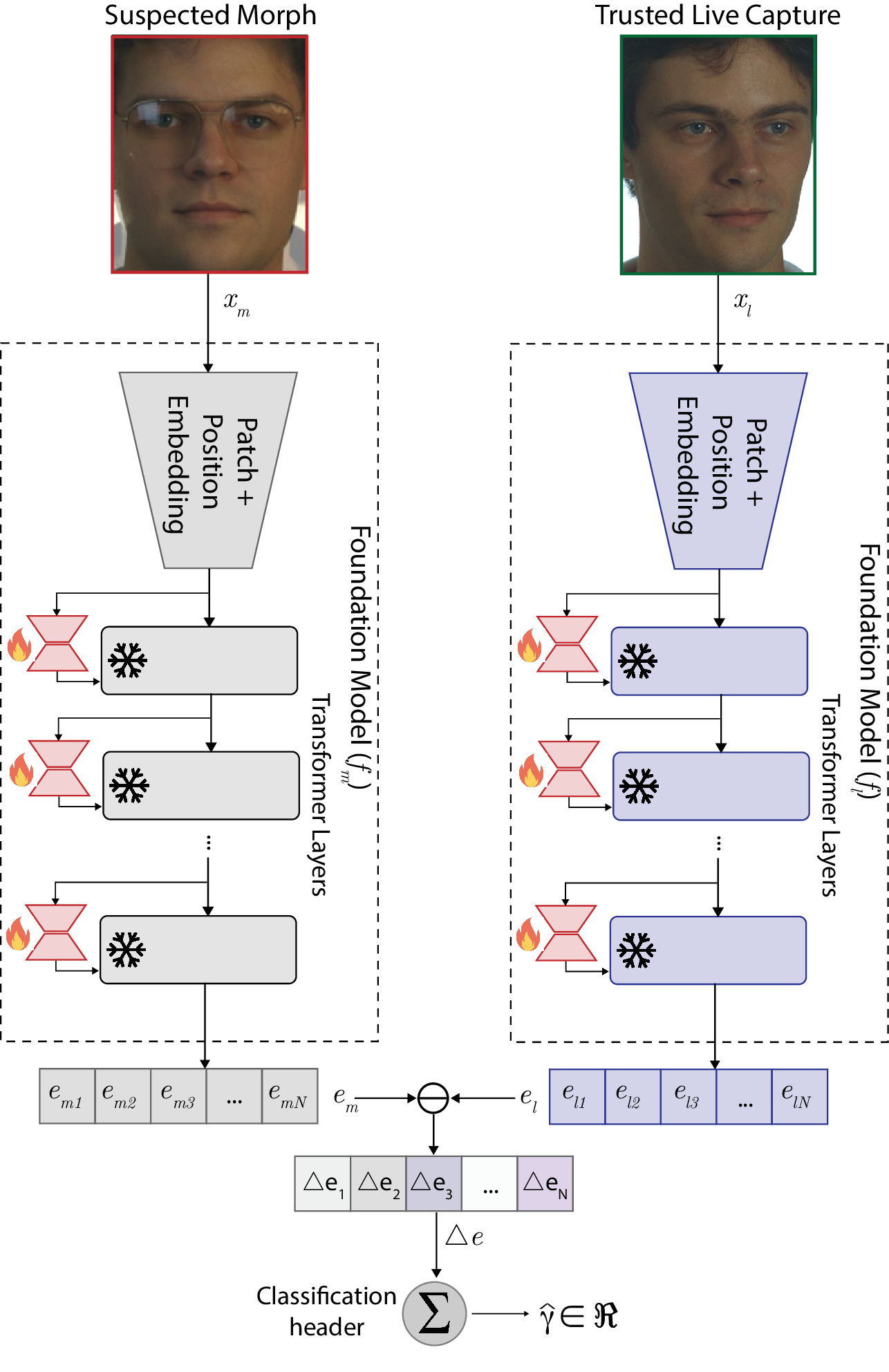}
    \caption{Overview of DifFoundMAD. Morphing artefacts are detected by computing a differential representation between embeddings from the suspected morph and the live capture using two adapted instances of the same FM. Fire and snowflake symbols denote trainable (LoRA) and frozen components, respectively.}
    \label{fig:overview}
\end{figure}

Existing D-MAD systems typically rely on a single feature space or identity-metric similarities. Such approaches implicitly assume that morph inconsistencies can be captured within identity embeddings. However, as modern morphing techniques aim to preserve identity similarity, the discriminative signals may become more subtle and potentially less aligned with identity representations. This suggests that identity embeddings alone may not always capture all relevant artefacts, particularly for advanced morphing pipelines.

DifFoundMAD follows the standard differential paradigm, but replaces identity representations with embeddings extracted from FMs. Instead of enforcing similarity constraints, the method models the discrepancy between a suspected morph and a trusted live capture directly in the embedding space, enabling the detection of fine-grained inconsistencies that may not be fully captured by identity representations.

Fig.~\ref{fig:overview} illustrates the proposed architecture. The system operates on paired inputs: a live capture and a suspected morph, which may correspond to a BS or a synthetically generated morph~\cite{Joshi-SyntheticDataSurvey-PAMI-2024}. In our work, we explore multiple FMs, including CLIP~\cite{Radford-CLIP-ICML-2021}, DINOv2~\cite{Oquab-DINOv2-ArXiv-2023}, DINOv3~\cite{Simeoni-DINOv3-ArXiv-2025}, and AIMv2~\cite{Fini-AImv2-CVPR-2025}. For CLIP, only the visual encoder is used.

DifFoundMAD employs a dual-stream architecture in which both inputs are processed by two instances of the same FM. Each branch is independently adapted using LoRA~\cite{Hu-LoRA-ICLR-2022}, while the pretrained backbone weights remain frozen. Although the architecture is shared, the weights are not shared, allowing each branch to specialise in encoding complementary characteristics of the live capture and the suspected morph. A single shared encoder applies identical feature transformations to both inputs, limiting the model to symmetric representations and reducing its ability to capture subtle, role-specific discrepancies. We also evaluated a shared-encoder variant in preliminary experiments, which resulted in degraded performance. In contrast, the proposed untied design enables asymmetric specialisation, where one branch focuses on stable identity cues from the live capture, while the other adapts to capture morph-specific artefacts and inconsistencies.

Let $f_{m}$ and $f_{l}$ denote the two adapted FM instances. Given a suspected morph $x_m$ and a live capture $x_l$, the corresponding embeddings are computed as:

\begin{equation}
    \mathbf{e}_m = f_{m}(x_m), \quad \mathbf{e}_l = f_{l}(x_l).
\end{equation}

The differential representation is then obtained by:
\begin{equation}
    \Delta \mathbf{e} = \mathbf{e}_l - \mathbf{e}_m,
\end{equation}
which captures discrepancies between the two inputs in the learned representation space.

This differential embedding is fed into a lightweight classification head that predicts the likelihood of a MA. Unlike identity-metric approaches, this formulation allows the model to learn subtle inconsistencies, including local texture deviations and structural artefacts, that are not explicitly encoded in identity embeddings.

During training, DifFoundMAD is optimised using Focal Loss (FL)~\cite{Lin-FocalLoss-ICCV-2017}, which down-weights easy samples and emphasises harder ones. This is particularly important in D-MAD, where datasets are often imbalanced and high-quality morphs are difficult to detect. The loss is defined as~\cite{Lin-FocalLoss-ICCV-2017}:

\begin{equation}
    \mathcal{L} = - \alpha_t (1 - p_t)^{\eta} \log(p_t),
\end{equation}

\noindent where \(\hat{\gamma}_i\) denotes the logit predicted for sample \(i\), \(\gamma_i \in \{0,1\}\) is the corresponding ground-truth label (1 for BSs and 0 for MAs), and \(\sigma(\cdot)\) is the sigmoid activation function. The term \(p_t\) represents the probability assigned to the ground-truth class, i.e., \(p_t=\sigma(\hat{\gamma}_i)\) for BSs and \(p_t=1-\sigma(\hat{\gamma}_i)\) for MAs. Here, \(\alpha_t\) balances class contributions, while \(\eta\) is the focusing parameter that down-weights easy samples and emphasises harder ones. The choice of FL is further motivated by previous PAD studies, where it demonstrated superior performance to Binary Cross-Entropy Loss under class imbalance and challenging attack conditions~\cite{George-CrossModalFocalLoss-CVPR-2021}.

By combining pretrained FM representations with parameter-efficient LoRA adaptation, DifFoundMAD effectively leverages large-scale visual priors while requiring only limited MAD-specific training data. The differential formulation further enhances robustness under cross-database and unknown-attack conditions.

\subsection{Low-Rank Adaptation (LoRA)}

FMs such as CLIP and DINO typically require large-scale datasets for full finetuning, as updating all parameters is computationally expensive and memory-intensive~\cite{Hu-LoRA-ICLR-2022}. When finetuned on small datasets, fully optimised models may override the knowledge learned from large and diverse pretraining data, leading to catastrophic forgetting. LoRA addresses this limitation by freezing pretrained weights and introducing trainable low-rank updates.

Formally, a weight matrix $W$ is decomposed as $W~=~W_0 + \Delta W$, where $W_0$ is the frozen pretrained weight and $\Delta W$ is the learnable update.

\begin{figure*}[!t]
    \centering
    \begin{subfigure}{0.95\linewidth}
        \includegraphics[width=\linewidth]{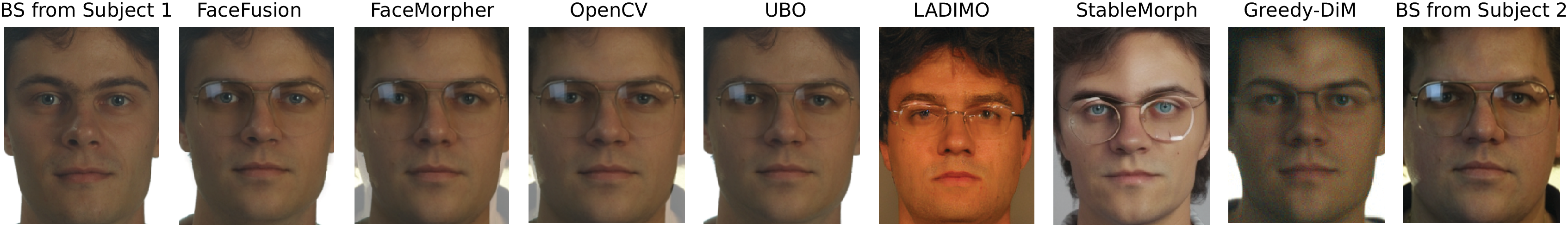}
        \caption{FERET}
    \end{subfigure} 
    \begin{subfigure}{0.95\linewidth}
        \includegraphics[width=\linewidth]{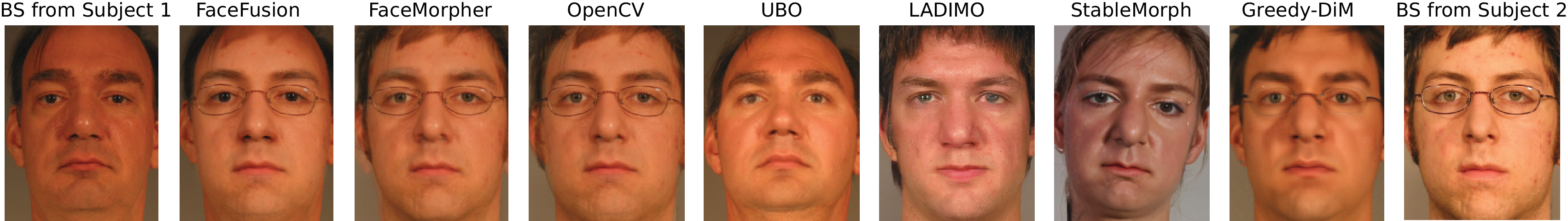}
        \caption{FRGC}
    \end{subfigure} 
    \caption{Example of BSs for two subjects contributing to different MAs for the FERET and FRGC databases.}
    \label{fig:databases}
\end{figure*}

Instead of learning $\Delta W$ directly, LoRA parameterises it as $\Delta W = B A$, where \(A \in \mathbb{R}^{r \times k}\), \(B \in \mathbb{R}^{d \times r}\), and \(r~\ll~\min(d,k)\). During forward propagation $W x~=~W_0 x + B A x.$

To improve training stability at higher ranks, we adopt Rank-Stabilised LoRA (rsLoRA)~\cite{Kalajdzievski-rsLoRA-ArXiv-2023}, which replaces the standard scaling factor \(\alpha / r\) with \(\alpha / \sqrt{r}\).

LoRA significantly reduces the number of trainable parameters (less than 1\%) and memory footprint while maintaining strong performance, making it particularly suitable for adapting large FMs to D-MAD with limited and less diverse data.
 
\section{Experimental Setup}
\label{sec:exp_setup}

The experimental evaluation assesses the generalisation capability of DifFoundMAD under realistic operational conditions. Our study pursues two goals: $(i)$ to analyse how different LoRA and optimisation settings affect the detection performance of DifFoundMAD; and $(ii)$ to benchmark DifFoundMAD against state-of-the-art D-MAD systems in operational scenarios. All experiments follow strict cross-database protocols, the most challenging setting for MAD, where training and testing datasets differ. We consider two variants: one where the morphing tools used for testing also appear in training (known as known-attack scenarios), and harder unknown-attack scenarios where test morphing tools are entirely unknown.

\subsection{Databases}
\label{sec:databases}

\begin{table}[!t]
	\centering
	\caption{A summary of existing databases used in D-MAD research or designed specifically for D-MAD research.}
	\label{tab:DB}
    \begin{adjustbox}{width=\linewidth}
	\begin{tabular}{r| c c c l} \toprule \toprule
            \textbf{DB}     &   \textbf{\#Images}     &     \textbf{\#BS} &  \textbf{\#MA}  &  \textbf{Morphing tools}     \\
\midrule
\midrule
\multirow{3}{*}{FERET~\cite{Phillips-FERET-1998}} & \multirow{3}{*}{5,024} 	  &         \multirow{3}{*}{1,321}      &  \multirow{3}{*}{3,703}       &	\multirow{6}{*}{\shortstack[l]{FaceFusion~\cite{Scherhag-FaceMorphingAttacks-TIFS-2020}, OpenCV~\cite{Scherhag-FaceMorphingAttacks-TIFS-2020}, \\ UBO-Morpher~\cite{Ferrara-FaceDemorphing-TIFS-2018}, \\ FaceMorpher~\cite{Scherhag-FaceMorphingAttacks-TIFS-2020}, \\ LADIMO~\cite{Grimmer-LADIMO-IJCB-2024}, \\ StableMorph~\cite{Kabbani-StableMorph-IJCB-2025}, and \\ Greedy-DiM~\cite{Blasingame-GreedyDiM-TBIOM-2024,Blasingame-GreedyDiM-IJCB-2024,Blasingame-FastDiM-SP-2024}}}    \\ 
	                   & 					   &	       &         &       \\ 
                      & 						&	        &         &       \\ 
\cmidrule{1-4}
\multirow{3}{*}{FRGCv2~\cite{Phillips-FRGC-CVPR-2005}} & \multirow{3}{*}{9,458} 	 &   \multirow{3}{*}{2,710} &  \multirow{3}{*}{6,748} &	 \\ 
	                   & 					  &	           &         &          \\ 
                      & 					   &	        &         &          \\ 
\midrule
\multirow{2}{*}{PMDB~\cite{Ferrara-FaceDemorphing-TIFS-2018}} & \multirow{2}{*}{1,380} 	 &       \multirow{2}{*}{280} &  \multirow{2}{*}{1,108} & \multirow{2}{*}{Landmark Morph~\cite{Ferrara-FaceDemorphing-TIFS-2018}}	 \\ 
	               & 						  &	                        &         &          \\ 
\midrule
\multirow{2}{*}{MorphDB~\cite{Ferrara-FaceDemorphing-TIFS-2018}} & \multirow{2}{*}{300} 	 &      \multirow{2}{*}{200} &  \multirow{2}{*}{100} &\multirow{2}{*}{Sqirlz Morph 2.1~\cite{Ferrara-FaceDemorphing-TIFS-2018}}	 \\ 
	               & 						  &	                      &         &          \\ 
\midrule
\multirow{2}{*}{Idiap Morph~\cite{Sarkar-IdiapMorph-ICASSP-2022}} & \multirow{2}{*}{1,020} 	 &       \multirow{2}{*}{408} &  \multirow{2}{*}{612} & \multirow{2}{*}{\shortstack[l]{StyleGAN~\cite{Karras-StyleGAN-CVPR-2020}, OpenCV~\cite{Scherhag-FaceMorphingAttacks-TIFS-2020}\\ and FaceMorpher~\cite{Scherhag-FaceMorphingAttacks-TIFS-2020}}}	  \\ 
	               & 						  &	                       &         &          \\ 
\midrule
\multirow{2}{*}{FEI Morph~\cite{Thomaz-FEI-JIVC-2010,Domenico-CombIdentityDMAD-ICIAP-2023}} & \multirow{2}{*}{6,200} 	 &     \multirow{2}{*}{200} &  \multirow{2}{*}{6,000} & \multirow{2}{*}{\shortstack[l]{FaceFusion~\cite{Scherhag-FaceMorphingAttacks-TIFS-2020}, UTW~\cite{Raja-MADBenchmark-TIFS-2020}\\ and NTNU~\cite{Raja-MADBenchmark-TIFS-2020}}}	 \\ 
	               & 						  &	                      &         &          \\ 
\midrule
\multirow{2}{*}{SynMorph~\cite{Zhang-SynMorph-Access-2025}} & \multirow{2}{*}{710K} 	 &     \multirow{2}{*}{480K} &  \multirow{2}{*}{230K} & \multirow{2}{*}{\shortstack[l]{MIPGAN-II~\cite{Zhang-MIPGAN-TBIOM-2021} and \\ UBO-Morpher~\cite{Ferrara-FaceDemorphing-TIFS-2018}}}	 \\ 
	               & 						  &	                       &         &          \\ 
\bottomrule \bottomrule
	 \end{tabular}
     \end{adjustbox}
\end{table}

To achieve these goals, the experimental evaluation is carried out on two publicly available databases for D-MAD, containing at least two ICAO-compliant facial images per subject and the widest variety of morphing tools: FERET~\cite{Phillips-FERET-1998} and FRGCv2~\cite{Phillips-FRGC-CVPR-2005} (see Fig.~\ref{fig:databases}). All FERET samples are taken in a controlled environment, but contain variations in pose and expression. FRGCv2 contains images suitable as passport photos, but also images with scene variations, e.g. non-uniform lighting, unsharpness and uneven background, suitable as live capture images. While FERET~\cite{Phillips-FERET-1998} consists of around 5,024 images split into 1,321 BSs and 3,703 MAs, FRGCv2~\cite{Phillips-FRGC-CVPR-2005} comprises a total of 9,458 images (2,710 BSs and 6,748 MAs). The MAs are derived from samples in the FERET and FRGCv2 datasets using four landmark-based morphing tools: FaceFusion~\cite{Scherhag-FaceMorphingAttacks-TIFS-2020}, FaceMorpher~\cite{Scherhag-FaceMorphingAttacks-TIFS-2020}, OpenCV~\cite{Scherhag-FaceMorphingAttacks-TIFS-2020}, and UBO-Morpher~\cite{Ferrara-FaceDemorphing-TIFS-2018}, as well as three diffusion-based models: LADIMO~\cite{Grimmer-LADIMO-IJCB-2024}, StableMorph~\cite{Kabbani-StableMorph-IJCB-2025}, and Greedy-DiM~\cite{Blasingame-GreedyDiM-TBIOM-2024,Blasingame-GreedyDiM-IJCB-2024,Blasingame-FastDiM-SP-2024}. FERET and FRGCv2 contain 529 and 964 MA samples per morphing tool, respectively. Tab.~\ref{tab:DB} summarises the main characteristics of the datasets.

\subsection{Implementation Details}

\begin{table}[t]
\centering
\caption{Training and LoRA hyperparameters used in DifFoundMAD.}
\label{tab:hyper}
\begin{adjustbox}{width=0.9\linewidth,center}
\begin{tabular}{l c}
\toprule
\textbf{Parameter} & \textbf{Value} \\
\midrule
Input size & $224 \times 224$ \\
Batch size & 32 (FERET), 64 (FRGC) \\
Training setup & 30 epochs, Adam, LR $1\mathrm{e}{-4}$, WD 0.01 \\
\midrule
LoRA & $r \in \{2,4,8\}$, $\alpha \in \{4,8,16\}$, $d \in \{0.2,0.4\}$ \\
LoRA layers & Q, V projections \\
\midrule
Focal Loss & $\alpha_t=0.25$, $\eta=2.0$ \\
\bottomrule
\end{tabular}
\end{adjustbox}
\end{table}

To comply with the input resolution of the FMs~\cite{Radford-CLIP-ICML-2021,Oquab-DINOv2-ArXiv-2023}, all images were resized to $224 \times 224$ pixels following~\cite{Caldeira-MADtion-WACV-2025}. Training included data augmentation (random cropping, horizontal flipping, and photometric transformations) and balanced sampling to maintain a 1:1 ratio between BSs and MAs~\cite{GonzalezSoler-ZeroFoundationModelsPAD-FG-2025}.

CLIP was initialised from LAION-400M~\cite{Schuhmann-LAION400M-ArXiv-2021}, while DINOv2 and DINOv3 were initialised from LVD-142M~\cite{Oquab-DINOv2-ArXiv-2023} and LVD-1689M~\cite{Simeoni-DINOv3-ArXiv-2025}, respectively. AIMv2 was initialised from its publicly available pretrained weights~\cite{Fini-AImv2-CVPR-2025}.

DifFoundMAD was implemented in PyTorch~\cite{Paszke-PyTorchAnImperative-2019} and trained for 30 epochs using Adam on an NVIDIA A100 GPU (80\,GB). LoRA was applied to the query (Q) and value (V) projections of the transformer blocks. Training each configuration required a few hours. The main hyperparameters are summarised in Tab.~\ref{tab:hyper}.

\begin{table*}[t!]
    \centering
    \caption{Cross-database performance (in \%) of DifFoundMAD for different LoRA parameters and FM architectures on known-attack scenarios. The best average performance per FM is highlighted in bold.}
    \label{tab:alpha_op_same}
    \begin{adjustbox}{width=0.8\linewidth,center}
    \begin{tabular}{c| c |c| c| c| c c c c| c c c c| c c c c } \toprule \toprule
           &  &  &   &   & \multicolumn{4}{c|}{\textbf{FERET $\rightarrow$ FRGC}} & \multicolumn{4}{c|}{\textbf{FRGC $\rightarrow$ FERET}} & \multicolumn{4}{c}{\textbf{Avg.}}    \\ \cmidrule{6-17}
            
 & & &  &  & \multirow{2}{*}{D-EER}  & \multicolumn{3}{c|}{BSCER@MACER} & \multirow{2}{*}{D-EER}  & \multicolumn{3}{c|}{BSCER@MACER} & \multirow{2}{*}{D-EER}  & \multicolumn{3}{c}{BSCER@MACER}     \\
\textbf{FM}   & \textbf{Backbone} &\textbf{$\mathbf{r}$} & \textbf{$\alpha$} & \textbf{$\mathbf{d}$} &     &   10\%  &  5\%  &  1\% &    &   10\%  &  5\%  &  1\% &    &   10\%  &  5\%  &  1\%    \\
\midrule
\midrule

  \multirow{6}{*}{AIMv2}  & \multirow{6}{*}{Vit-L/14}  & \multirow{2}{*}{2}  &  \multirow{2}{*}{4}    &  0.2  & 7.25 & 4.45 & 11.18 & 40.22 & 1.40 & 0.56 & 0.98 & 2.11 & 4.33 & 2.50 & 6.08 & 21.16 \\
 & &          &         &  0.4  & 7.08 & 4.55 & 10.04 & 35.97 & 1.69 & 0.56 & 0.70 & 2.53 & 4.39 & 2.55 & 5.37 & 19.25 \\  \cmidrule{3-17}
 & & \multirow{2}{*}{4}  &  \multirow{2}{*}{8}    &  0.2  & 4.31 & 1.18 & 3.84 & 18.39 & 1.69 & 0.42 & 0.98 & 2.25 & \textbf{3.00} & \textbf{0.80} & \textbf{2.41} & \textbf{10.32} \\
 & &                     &                        &  0.4  & 4.42 & 1.31 & 3.64 & 18.63 & 2.23 & 0.42 & 1.27 & 3.94 & 3.33 & 0.87 & 2.45 & 11.28 \\
 \cmidrule{3-17}
 & & \multirow{2}{*}{8}  &  \multirow{2}{*}{16}   &  0.2  & 4.08 & 1.08 & 3.27 & 18.19 & 2.23 & 0.28 & 0.84 & 4.08 & 3.16 & 0.68 & 2.06 & 11.13 \\
 & &                     &                        &  0.4  & 4.41 & 1.15 & 3.67 & 23.75 & 2.23 & 0.42 & 0.70 & 4.08 & 3.32 & 0.78 & 2.19 & 13.91 \\

 \midrule

\multirow{6}{*}{CLIP}  & \multirow{6}{*}{Vit-L/14}  & \multirow{2}{*}{2}  &  \multirow{2}{*}{4}    &  0.2  & 1.75 & 0.10 & 0.34 & 2.53 & 5.27 & 2.67 & 5.34 & 18.14 & 3.51 & 1.39 & 2.84 & 10.33 \\
 & &          &         &  0.4  & 2.07 & 0.07 & 0.44 & 4.14 & 6.90 & 4.78 & 8.58 & 26.86 & 4.48 & 2.42 & 4.51 & 15.50 \\ \cmidrule{3-17}
 & & \multirow{2}{*}{4}  &  \multirow{2}{*}{8}    &  0.2  & 0.94 & 0.03 & 0.03 & 0.91 & 3.23 & 0.84 & 1.97 & 10.13 & \textbf{2.09} & \textbf{0.44} & \textbf{1.00} & \textbf{5.52} \\
 & &                     &                        &  0.4  & 0.65 & 0.00 & 0.03 & 0.37 & 3.11 & 0.56 & 1.41 & 12.52 & 1.88 & 0.28 & 0.72 & 6.44 \\
 \cmidrule{3-17}
 & & \multirow{2}{*}{8}  &  \multirow{2}{*}{16}   &  0.2  & 0.57 & 0.00 & 0.00 & 0.40 & 6.93 & 5.34 & 9.85 & 25.60 & 3.75 & 2.67 & 4.92 & 13.00 \\
 & &                     &                        &  0.4  & 0.72 & 0.00 & 0.03 & 0.37 & 4.91 & 1.97 & 4.92 & 14.35 & 2.81 & 0.98 & 2.48 & 7.36 \\
    \midrule

\multirow{6}{*}{DINOv2}  & \multirow{6}{*}{Vit-L/14}  & \multirow{2}{*}{2}  &  \multirow{2}{*}{4}    &  0.2  & 1.08 & 0.10 & 0.24 & 1.15 & 4.92 & 2.81 & 4.64 & 19.97 & 3.00 & 1.46 & 2.44 & 10.56  \\
 & &          &         &  0.4  & 1.24 & 0.07 & 0.27 & 1.55 & 3.70 & 0.84 & 1.55 & 14.06 & 2.47 & 0.46 & 0.91 & 7.81 \\ \cmidrule{3-17}
 & & \multirow{2}{*}{4}  &  \multirow{2}{*}{8}    &  0.2  & 0.87 & 0.03 & 0.10 & 0.84 & 1.82 & 0.14 & 0.56 & 3.66 & \textbf{1.34} & \textbf{0.09} & \textbf{0.33} & \textbf{2.25} \\
 & &                     &                        &  0.4  & 0.91 & 0.10 & 0.10 & 0.84 & 3.37 & 1.69 & 2.39 & 8.02 & 2.14 & 0.89 & 1.25 & 4.43 \\
 \cmidrule{3-17}
 & & \multirow{2}{*}{8}  &  \multirow{2}{*}{16}   &  0.2  & 0.81 & 0.00 & 0.10 & 0.67 & 4.00 & 1.83 & 3.23 & 11.81 & 2.40 & 0.91 & 1.67 & 6.24 \\
 & &                     &                        &  0.4  & 0.58 & 0.00 & 0.10 & 0.37 & 4.07 & 1.69 & 3.23 & 16.60 & 2.32 & 0.84 & 1.67 & 8.48 \\
    \midrule

 \multirow{6}{*}{DINOv3}  & \multirow{6}{*}{Vit-L-16}  & \multirow{2}{*}{2}  &  \multirow{2}{*}{4}    &  0.2  & 1.66 & 0.00 & 0.17 & 3.20 & 2.95 & 1.13 & 1.97 & 6.61 & 2.30 & 0.56 & 1.07 & 4.91 \\
 & &          &         &  0.4  & 1.16 & 0.00 & 0.00 & 1.68 & 3.37 & 1.27 & 2.11 & 7.03 & 2.26 & 0.63 & 1.05 & 4.36 \\ \cmidrule{3-17}
 & & \multirow{2}{*}{4}  &  \multirow{2}{*}{8}    &  0.2  & 1.31 & 0.07 & 0.07 & 1.62 & 2.83 & 0.98 & 1.69 & 6.33 & 2.07 & 0.53 & 0.88 & 3.97 \\
 & &                     &                        &  0.4  & 1.14 & 0.00 & 0.07 & 1.18 & 3.37 & 1.13 & 3.09 & 9.00 & 2.26 & 0.56 & 1.58 & 5.09 \\
 \cmidrule{3-17}
 & & \multirow{2}{*}{8}  &  \multirow{2}{*}{16}   &  0.2  & 0.71 & 0.00 & 0.00 & 0.27 & 2.13 & 0.00 & 0.00 & 4.08 & \textbf{1.42} & \textbf{0.00} & \textbf{0.00} & \textbf{2.17} \\
 & &                     &                        &  0.4  & 0.94 & 0.00 & 0.00 & 0.94 & 3.11 & 0.56 & 1.69 & 5.77 & 2.02 & 0.28 & 0.84 & 3.35 \\
        \bottomrule
       \bottomrule
    \end{tabular}
    \end{adjustbox}
\end{table*}

\subsection{Evaluation Metrics}






The experimental results are reported following ISO/IEC 20059~\cite{ISO-IEC-20059} for biometric MAD. We consider $i)$ Morphing Attack Classification Error Rate (MACER), which measures the proportion of MAs misclassified as BSs, and $ii)$ Bona fide Sample Classification Error Rate (BSCER), which quantifies the proportion of BSs misclassified as MAs.

These metrics are particularly relevant in operational scenarios such as border control and automated eGate systems, where accepting a MA is significantly more critical than rejecting a bona fide traveller. Therefore, we report BSCER at fixed MACER thresholds of 1\% (BSCER100), 5\% (BSCER20), and 10\% (BSCER10), corresponding to strict security requirements, as well as the Detection Equal Error Rate (D-EER), defined at the operating point where MACER equals BSCER.

\section{Results and Discussion}
\label{sec:results}

\subsection{Parameter Optimisation}
\label{sec:po_similar_FM}

A comprehensive benchmark across all parameter configurations in Tab.~\ref{tab:alpha_op_same}, which lists only the best-performing ViT architecture for each FM, shows that DifFoundMAD is strongly influenced by both the selection of FM and the LoRA parameters. Note that, in this evaluation, the same landmark-based morphing tools are used in both training and testing, thereby isolating the effect of LoRA parameter selection from cross-tool generalisation. Across all models, moderate ranks ($\mathbf{r}=4$) and scaling factors ($\alpha=8$) provide the best trade-off between adaptation capacity and generalisation. Lower ranks (e.g., $\mathbf{r}=2$) tend to underfit, while higher ranks (e.g., $\mathbf{r}=8$) often degrade cross-database performance. Similarly, a dropout of $\mathbf{d}=0.2$ consistently yields more stable results than $\mathbf{d}=0.4$.

Among all FMs, DINO-based architectures achieve the strongest performance. In particular, DINOv2 (ViT-L/14) with $\mathbf{r}=4$, $\alpha=8$, and $\mathbf{d}=0.2$ achieves the best overall results, reaching an average D-EER of $1.34\%$ and BSCER values of $0.09\%$, $0.33\%$, and $2.25\%$ at MACER thresholds of $10\%$, $5\%$, and $1\%$, respectively. DINOv3 demonstrates comparable performance, achieving an average D-EER of $1.42\%$, and shows particularly favourable behaviour at strict security levels, with near-zero BSCER values at higher security thresholds (i.e., BSCER10=0.0\%, BSCER20=0.0\% and BSCER100=2.17\%).

Note that CLIP-based DifFoundMAD exhibits less stable behaviour. While it achieves very low error rates in the FERET $\rightarrow$ FRGC scenario, its performance degrades notably in the reverse direction, leading to higher average errors and reduced robustness under cross-database shifts. In contrast, AIMv2 performs significantly worse across all configurations, indicating that not all FMs are equally suitable for D-MAD.

The superior performance of DINO-based models may be attributed to their self-supervised training paradigm, which has been shown to preserve fine-grained spatial and structural information. While this behaviour is not explicitly analysed in this work, our results are consistent with this property, particularly at strict security levels. In contrast, CLIP is optimised for image--text alignment and tends to favour invariance to local variations, reducing sensitivity to the subtle artefacts required for effective D-MAD. 


\subsection{Benchmark against the State-of-the-Art}

\begin{table*}[!t]
    \centering
    \caption{Benchmark (in \%) of DifFoundMAD with the state-of-the-art for cross-database and known-attack scenarios using the best average parameter combination from Tab.~\ref{tab:alpha_op_same}.}
    \label{tab:benchmark}
    \begin{adjustbox}{width=0.95\linewidth,center}
    \begin{threeparttable}
    \begin{tabular}{r|| c|c|c|c||c|c|c|c||c|c|c|c}
    \toprule \toprule
    \textbf{Method}  & \multicolumn{4}{c||}{\textbf{FERET $\rightarrow$ FRGC}} & \multicolumn{4}{c||}{\textbf{FRGC $\rightarrow$ FERET}} & \multicolumn{4}{c}{\textbf{Avg.}} \\ 
              & D-EER & BSCER10 & BSCER20 & BSCER100 & D-EER & BSCER10 & BSCER20 & BSCER100 & D-EER & BSCER10 & BSCER20 & BSCER100 \\ 
    \midrule
    MagFace+SVM~\cite{Kessler-MinPairSelectionMAD-2024} & 3.06 & 0.55 & 1.64 & 10.08 & 1.60 & 0.00 & 0.12 & \textbf{2.24} & 2.33 & 0.28 & 0.88 & 6.16\\ 
    
    ArcFace+SVM~\cite{Scherhag-FaceMorphingAttacks-TIFS-2020} & 4.90 & 2.07 & 4.87 & 18.39 & 1.91 & 0.27 & 0.80 & 3.19 & 3.41 & 1.17 & 2.84 & 10.79\\ 
    ArcFace+RF~\cite{Scherhag-FaceMorphingAttacks-TIFS-2020} & 10.85 & 11.60 & 19.21 & 36.57 & 2.52 & 0.66 & 1.46 & 5.71 & 6.69 & 6.13 & 10.34 & 21.14 \\ 
    FaceNet+SVM~\cite{Scherhag-FaceMorphingAttacks-TIFS-2020} & 23.48 & 46.24 & 60.69 & 85.25 & 12.01 & 13.69 & 21.38 & 44.72 & 17.75 & 29.97 & 41.04 & 64.99\\ 
    FaceNet+RF~\cite{Scherhag-FaceMorphingAttacks-TIFS-2020} & 31.64 & 60.05 & 72.16 & 87.74 & 13.02 & 15.91 & 20.47 & 43.94 & 22.33 & 37.98 & 46.32 & 65.84\\ 
    COTS$^{*}$+SVM~\cite{Scherhag-FaceMorphingAttacks-TIFS-2020} & 14.27 & 19.10 & 30.15 & 60.29 & 4.95 & 3.26 & 4.82 & 13.82 & 9.61 & 11.18 & 17.49 & 37.06\\ 
    COTS$^{*}$+RF~\cite{Scherhag-FaceMorphingAttacks-TIFS-2020} & 19.55 & 33.12 & 49.91 & 72.07 & 9.66 & 9.65 & 14.08 & 28.55 & 14.61 & 21.39 & 32.00 & 50.31\\ 
    \midrule \midrule
    DifFoundMAD[AIMv2(Vit-L/14)] & 4.31 & 1.18 & 3.84 & 18.39 & \textbf{1.69} & 0.42 & 0.98 & 2.25 & 3.00 & 0.80 & 2.41 & 10.32 \\
    DifFoundMAD[CLIP(Vit-L/14)] & 0.94 & 0.03 & 0.03 & 0.91 & 3.23 & 0.84 & 1.97 & 10.13 & 2.09 & 0.44 & 1.00 & 5.52 \\
    DifFoundMAD[DINOv2(Vit-L/14)] & 0.87 & 0.03 & 0.10 & 0.84 & 1.82 & 0.14 & 0.56 & 3.66 & \textbf{1.34} & 0.09 & 0.33 & 2.25  \\
    DifFoundMAD[DINOv3(Vit-L/14)] & \textbf{0.71} & \textbf{0.00} & \textbf{0.00} & \textbf{0.27} & 2.13 & \textbf{0.00} & \textbf{0.00} & 4.08 & 1.42 & \textbf{0.00} & \textbf{0.00} & \textbf{2.17} \\
    \bottomrule \bottomrule
    \end{tabular}
    \begin{tablenotes}\footnotesize
		\item[*] Eyedea: \url{https://www.eyedea.ai/eyeface-sdk/}
    \end{tablenotes}
    \end{threeparttable}
    \end{adjustbox}
\end{table*}

Using the best-performing LoRA parameters per FM, we benchmark DifFoundMAD against the state-of-the-art in Tab.~\ref{tab:benchmark}. As in the previous experiment in Sect.~\ref{sec:po_similar_FM}, the same landmark-based morphing tools are used for both training and evaluation. Note that DifFoundMAD consistently outperforms traditional face recognition-based approaches in cross-database scenarios. Existing D-MAD methods built on the latent embeddings of ArcFace, FaceNet, and Commercial Off-The-Shelf (COTS), exhibit significantly higher error rates, with average D-EER values ranging from 6.69\% to 22.33\%, and BSCER100 often exceeding 21--65\%. Although MagFace+SVM is the strongest baseline (D-EER = 2.33\%), its performance is highly asymmetric, degrading notably for FERET$\rightarrow$FRGC and it still underperforms DifFoundMAD for CLIP and DINO. This performance gap suggests that FM representations capture subtle inconsistencies that may not be fully exploited by identity-based D-MAD systems.

In contrast, DifFoundMAD achieves superior and more balanced performance across both directions. DINO-based configurations provide the best results, with DINOv2 (ViT-L/14) achieving the lowest average D-EER (1.34\%) and DINOv3 (ViT-L/14) yielding the best performance at strict operating points, with near-zero BSCER and the lowest average BSCER100 of 2.17\%. CLIP also improves over traditional methods but shows less stable cross-database behaviour, while AIMv2 performs worse overall. In general, these results demonstrate that DifFoundMAD not only improves average detection performance but, more importantly, provides substantially lower error rates at strict security levels. This is particularly critical for real-world applications such as border control, where low false acceptance rates are required. The results further highlight that FMs preserving fine-grained structural information, such as DINO-based architectures, are best suited for capturing morphing artefacts under cross-database conditions. 

\subsection{Ablation Study}
\label{sec:ablation_study}

\begin{figure}[!t]
\centering
    \includegraphics[width=0.75\linewidth]{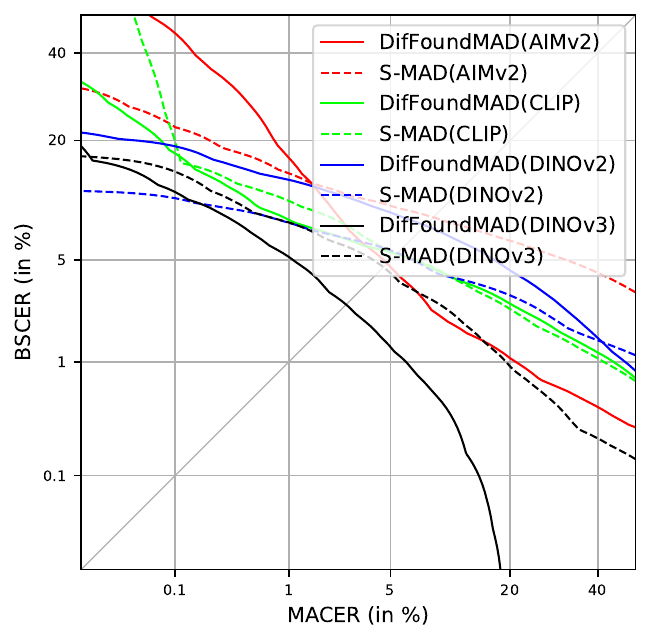}
    \caption{DET curves comparing DifFoundMAD and the corresponding S-MAD configurations for each FM. Each method is evaluated using its respective best-performing LoRA parameters selected from the search space in Tab.~\ref{tab:alpha_op_same}.}
    \label{fig:ablation}
\end{figure}

To analyse the contribution of the differential formulation, we repeat the experiments from Sect.~\ref{sec:po_similar_FM} using each FM in a single-image setting (i.e., S-MAD), where only the suspected morph is analysed. For a fair comparison, S-MAD models are optimised using the same LoRA search space. Note that the optimal LoRA parameters may differ between the S-MAD and differential formulations; therefore, in both cases, we report the performance using their respective best-performing configurations. The results, shown as DET curves in Fig.~\ref{fig:ablation}, reveal model-dependent behaviour.

Note that DifFoundMAD consistently outperforms S-MAD for DINOv3 and CLIP across all operating points, and also improves AIMv2 at moderate operating points (MACER $> 1\%$). Notably, DifFoundMAD with DINOv3 is the only configuration that enables a secure and convenient D-MAD subsystem, achieving both MACER and BSCER below 20\% simultaneously. This confirms that the differential formulation effectively exploits complementary information between the live capture and the suspected morph.

In contrast, DINOv2 performs slightly better in the S-MAD setting. This can be explained by the nature of the evaluated morphs, which are landmark-based and exhibit visible artefacts that DINOv2 can already capture in the S-MAD setting. In this case, the differential operation may introduce additional variability without providing significant new information. However, this advantage is expected to diminish for diffusion-based morphs, where artefacts are largely suppressed. In such scenarios, S-MAD approaches are likely to degrade, while the differential formulation remains effective by modelling inconsistencies between the live capture and the suspected morph. 

In general, the results highlight that the benefit of DifFoundMAD depends on the interaction between the representation and the differential formulation, and becomes particularly relevant as morphing techniques evolve towards more realistic generation.


\subsection{Benchmark on Unknown Attacks}

\begin{table}[!t]
\centering
\caption{Benchmark (in \%) of DifFoundMAD[DINOv3] against the state-of-the-art for unknown-attack scenarios. ``Landmark'' refers to training on landmark-based morphs, while ``Diffusion'' follows a leave-one-out protocol using two diffusion-based morphs for training and the rest for evaluation.}
\label{tab:cross_db_morphing}
\begin{adjustbox}{width=\linewidth}
\begin{tabular}{c|c|c||c c| c c}
\toprule \toprule
\multirow{3}{*}{\textbf{Train}}               & \multirow{2}{*}{\textbf{Test}} & \multirow{3}{*}{\textbf{Metrics}} &    \multicolumn{4}{c}{\textbf{Approaches}} \\
 &            &                  &  \multicolumn{2}{c|}{DifFoundMAD} & \multicolumn{2}{c}{MagFace+SVM~\cite{Kessler-MinPairSelectionMAD-2024}} \\
 &    FRGC    &                  &  Landmark  & Diffusion & Landmark  & Diffusion \\
\midrule
\multirow{12}{*}{FERET} & \multirow{4}{*}{LADIMO}      &  D-EER         & 4.43     &    \textbf{0.00}  & 1.24 & 0.95\\
                       &                               & BSCER10        & 0.47     &    \textbf{0.00}  & 0.06 & 0.06\\
                       &                               & BSCER20        & 3.84     &    \textbf{0.00}  & 0.15 & 0.12\\
                       &                               & BSCER100       & 26.00    &    \textbf{0.00}  & 1.49 & 0.82\\
\cmidrule{2-7}
                       & \multirow{4}{*}{StableMorph}  & D-EER          & 9.68     &    29.31 & 1.59 & \textbf{1.10} \\
                       &                               & BSCER10        & 8.79     &    67.40 & 0.09 & \textbf{0.06}\\
                       &                               & BSCER20        & 19.97    &    86.06 & 0.33 & \textbf{0.09}\\
                       &                               & BSCER100       & 55.61    &    97.34 & 2.73 & \textbf{1.28}\\
\cmidrule{2-7}
                       & \multirow{4}{*}{Greedy-DiM}   & D-EER          & 0.02     &    \textbf{0.00}  & 1.72 & 1.75\\
                       &                               & BSCER10        & \textbf{0.00}     &    \textbf{0.00}  & 0.21 & 0.18\\
                       &                               & BSCER20        & \textbf{0.00}     &    \textbf{0.00}  & 0.52 & 0.43\\
                       &                               & BSCER100       & \textbf{0.00}     &    \textbf{0.00}  & 3.04 & 2.55\\
\midrule \midrule
                        &    FERET & \cellcolor{gray!15} & \cellcolor{gray!15}  & \cellcolor{gray!15} & \cellcolor{gray!15}  & \cellcolor{gray!15}           \\
\midrule
\multirow{12}{*}{FRGC} & \multirow{4}{*}{LADIMO}       & D-EER          & 21.85    &    24.70 & 0.19 & \textbf{0.12}\\
                       &                               & BSCER10        & 36.57    &    40.51 & \textbf{0.00} & \textbf{0.00}\\
                       &                               & BSCER20        & 47.82    &    56.54 & \textbf{0.00} & \textbf{0.00}\\
                       &                               & BSCER100       & 74.26    &    87.06 & \textbf{0.00} & \textbf{0.00}\\
\cmidrule{2-7}
                       & \multirow{4}{*}{StableMorph}  & D-EER          & 29.55     &   0.56 & 0.64 & \textbf{0.32}\\
                       &                               & BSCER10        & 56.40     &   \textbf{0.00} & \textbf{0.00} & \textbf{0.00}\\
                       &                               & BSCER20        & 68.64     &   \textbf{0.00} & \textbf{0.00} & \textbf{0.00}\\
                       &                               & BSCER100       & 84.67     &   \textbf{0.00} & 0.25 & \textbf{0.00} \\
\cmidrule{2-7}
                       & \multirow{4}{*}{Greedy-DiM}   & D-EER          & 9.32      &   11.55 & 0.76 & 0.89\\
                       &                               & BSCER10        & 8.44      &   12.94 & \textbf{0.00} & \textbf{0.00}\\
                       &                               & BSCER20        & 17.86     &   17.58 & \textbf{0.00} & \textbf{0.00}\\
                       &                               & BSCER100       & 30.24     &   36.85 & 0.25 & 0.62\\
\bottomrule \bottomrule
\end{tabular}
\end{adjustbox}
\end{table}

To assess robustness under realistic deployment conditions, we evaluate DifFoundMAD[DINOv3 (ViT-L/14)] against unknown morphing tools using a leave-one-out protocol, where two diffusion-based morphs are used for training and the remaining one for testing. In addition, we evaluate each diffusion-based morph when both DifFoundMAD and MagFace+SVM are trained exclusively on landmark-based morphs, enabling the analysis of generalisation under strong domain shifts.

The results in Tab.~\ref{tab:cross_db_morphing} reveal performance behaviour that depends on the model and the attack type. DifFoundMAD achieves near-perfect detection performance when training and testing distributions are aligned, particularly for Greedy-DiM, where error rates are close to zero across both protocols. This is notable given that Greedy-DiM is considered a highly challenging attack due to its strong identity preservation~\cite{Blasingame-GreedyDiM-IJCB-2024} (morphing attack potential-MAP=100\% at FMR=0.1\%). Improvements are also observed for LADIMO under diffusion-based training, which operates in latent space and preserves identity information.


In contrast, detection performance degrades for StableMorph when trained on landmark-based data, reflecting limited generalisation under strong domain shifts (e.g., D-EER of 9.68\% and 29.55\% for FERET$\rightarrow$FRGC and FRGC$\rightarrow$FERET, respectively). When trained on diffusion-based morphs, performance improves significantly in FRGC$\rightarrow$FERET (D-EER drops to 0.56\%), while remaining challenging in FERET$\rightarrow$FRGC (D-EER of 29.31\%). This highlights the importance of aligning the training distribution with the target attack domain, as well as the increased difficulty of certain cross-database diffusion scenarios.

A consistent trend is that FERET, despite being smaller than FRGC, appears to provide more diverse training conditions, leading to better generalisation. MagFace+SVM shows more stable behaviour across conditions due to its reliance on identity-based comparison, but lacks sensitivity to subtle artefacts. These results indicate that while the differential formulation is effective, its performance depends on the training distribution. One promising approach involves combining identity-based cues with FM representations to improve resistance to realistic, artefact-free MAs.




\section{Conclusion}
\label{sec:conclusions}

 We presented DifFoundMAD, a D-MAD framework that leverages FM representations within a differential formulation to capture subtle inconsistencies between a live capture and a suspected morph. The proposed approach consistently outperforms traditional face recognition-based methods and remains competitive with strong identity-based baselines, with DINO-based models achieving the best overall performance, particularly at strict security levels. Our analyses show that performance depends on both the representation and the training distribution, and that generalisation to unseen diffusion-based morphs remains challenging under strong domain shifts. From an operational perspective, FMs introduce practical limitations, including increased model size, higher inference latency, and adaptation challenges. In particular, models such as CLIP, not specifically designed for facial analysis, may exhibit suboptimal behaviour in biometric scenarios. These constraints highlight the need for more efficient solutions.


Future work will focus on evaluating DifFoundMAD on NIST FATE MORPH and BOEP, while exploring hybrid identity--FM representations and lightweight models for real-time applications.

\section*{Acknowledgements}

This research work has been partially funded by the European Union (EU) under G.A. no. 101121280 (EINSTEIN) and CarMen (101168325), and UKRI Funding Service under IFS reference 10093453, and the German Federal Ministry of Education and Research and the Hessian Ministry of Higher Education, Research, Science and the Arts within their joint support of the National Research Center for Applied Cybersecurity ATHENE.





{\small
\bibliographystyle{ieee}
\bibliography{egbib}
}

\end{document}